\documentclass[3p,twocolumn]{elsarticle}
\usepackage{hyperref}
\journal{Journal of \LaTeX\ Templates}
\usepackage{amssymb}
\usepackage{graphicx}
\usepackage{textcomp}
\usepackage{amsmath,amssymb,amsfonts}
\usepackage{algorithmic}
\usepackage{graphicx}
\usepackage{caption}
\usepackage{textcomp}
\usepackage{color}
\usepackage[dvipsnames]{xcolor}
\usepackage{subfigure}
\usepackage{multirow}
\usepackage{rotating}
\usepackage{diagbox}
\usepackage{stfloats}
\usepackage{float}
\usepackage{bm}
\usepackage{booktabs}
\usepackage{float}
\usepackage{multirow}
\usepackage{soul}
\soulregister\cite7
\soulregister\citep7
\soulregister\citet7
\soulregister\ref7 
\soulregister\pageref7 
\usepackage{appendix}
\usepackage{gensymb}
\usepackage{float} 

\pdfoutput=1
\begin{document}

\begin{frontmatter}

\title{An Application of Pixel Interval Down-sampling (PID) for Dense Tiny Microorganism Counting on Environmental Microorganism Images}

\author[1address]{Jiawei Zhang}
\author[2address]{Xin Zhao}
\author[5address]{Tao Jiang}
\author[1address]{Md Mamunur Rahaman}
\author[3address]{Yudong Yao}
\author[4address]{Yu-Hao Lin}
\author[1address]{Jinghua Zhang}
\author[7address]{Ao Pan}
\author[6address]{Marcin Grzegorzek}
\author[1address]{Chen Li\corref{mycorrespondingauthor}}
\cortext[mycorrespondingauthor]{Corresponding author}
\ead{lichen201096@hotmail.com}

\address[1address]{Microscopic Image and Medical Image Analysis Group, College of 
Medicine and Biological Information Engineering, Northeastern University, Shenyang, China} 
\address[2address]{School of Resources and Civil Engineering, Northeastern University, Shenyang, China} 
\address[3address]{Stevens Institute of Technology, US}
\address[4address]{Department of Environmental Engineering, National Chung Hsing University, 145 Xingda Rd., Taichung 402, Taiwan}
\address[5address]{School of Control Engineering, Chengdu University of Information Technology, P.R. China}
\address[6address]{Institute of Medical Informatics, University of Luebeck,  Germany}
\address[7address]{School of Life Science and Technology, Huazhong University of Science and Technology, 430074, Wuhan, China}
\begin{abstract}
This paper proposes a novel pixel interval down-sampling network (PID-Net) for dense tiny object (yeast cells) counting tasks with higher accuracy.
The PID-Net is an end-to-end convolutional neural network (CNN) model with an encoder--decoder architecture.
The pixel interval down-sampling operations are concatenated with max-pooling operations to combine the sparse and dense features.
This addresses the limitation of contour conglutination of dense objects while counting.
The evaluation was conducted using classical segmentation metrics (the Dice, Jaccard and Hausdorff distance) as well as counting metrics.
The experimental results show that the proposed PID-Net had the best performance and potential for dense tiny object counting tasks, which achieved 96.97\% counting accuracy on the dataset with 2448 yeast cell images.
By comparing with the state-of-the-art approaches, such as Attention U-Net, Swin U-Net and Trans U-Net, the proposed PID-Net can segment  dense tiny objects with clearer boundaries and fewer incorrect debris, which shows the great potential of PID-Net in the task of accurate counting.
\end{abstract}


\end{frontmatter}

\section{Introduction}

With the development of industrialization, environmental pollution has become a vital problem to be resolved urgently. 
Compared with the classical physical and chemical approaches, the~novel biological methods are more efficient and transparent, causing no secondary pollution, which has become the preference for  environmental pollution.
The research of \emph{Environmental Microorganisms} 
 (EMs) is helpful to focus on the interrelationship among microorganisms, pollutants and~the environment. It is essential to use microorganisms to degrade the increasingly severe and diverse environmental pollutants effectively.

Yeast is a kind of single-celled eukaryotic microorganism that  is highly adaptable to the environment. 
It is widely applied to produce alcohol, glycerol and~organic acids, which are closely linked to the life and production activity of humanity.
Until now, yeast have also been used in the treatment of toxic industrial wastewater and solid waste, which plays an important role in treating environmental pollution~\cite{Yu-2020-ICPA,Wang-2018-AYTW}.

In the research of yeast applied in industrial production and environmental pollution control, biomass is the basic evaluation method and can quantitatively consider the performance of yeast in various tasks~\cite{You-2021-DAFD}.
At present, there are mainly two types of counting methods. The~first  is manual counting methods, such as plate counting and hemocytometry; another  is semi-automatic counting methods, such as flow cytometry~\cite{Balestra-1997-ITET,Sambrook-2006-ECNH}.

Manual counting is straightforward and stable to use with high accuracy when the number of cells is limited.
However, when the number of cells becomes larger, it will be time-consuming, and~the accuracy will be lowered due to the subjective influence of the operator. 
Semi-automatic counting is more accurate and can obtain ideal results in the case of large biomass; however,~it is not portable and requires expensive equipment~\cite{Gasol-2000-UFCC}. 
Therefore, these classical methods have non-negligible limitations in~practice.

Due to the rapid developments of computer vision and deep learning technologies, computer-assisted image analysis is broadly applied in many research fields, including histopathological image analysis~\cite{Li-2022-ACRC,Li-2021-ACRM,Zhou-2020-ACRB,Li-2020-ARCH}, cytopathological image analysis~\cite{Liu-2022-ITAT,Rahaman-2021-DADL,Rahaman-2020-ASCC}, object detection~\cite{Zou-2022-TAEC,Chen-2022-SDAN,Ma-2022-ASSO}, microorganism classification~\cite{Zhao-2022-ACAD,Kulwa-2022-ANPD,Kosov-2018-EMCC,Li-2016-EMAC,Li-2015-ACIA,Rahaman-2020-ICSC}, microorganism \mbox{segmentation~\cite{Zhang-2022-AANN,Kulwa-2019-ASSM,Zhao-2022-EEMI,Li-2020-ASMI}} and microorganism counting~\cite{Zhang-2022-AAPI,Li-2021-ACRI}.

However, by~reviewing the works of microorganism counting from the 1980s until now~\cite{Li-2021-ACRI}, we find that, in~the process of image segmentation, all existing segmentation approaches use traditional technologies, such as thresholding~\cite{Chunhachart-2016-CAVE}, edge detection~\cite{Choudhry-2016-HTMA} and watershed~\cite{Minoi-2016-MVBA}. 
Most of the deep-learning approaches are only applied for microorganism classification but~not for microorganism segmentation in the task of microorganism counting~\cite{Li-2019-ASTA}.
Here, we propose a novel \emph{Pixel Interval Down-sampling Network} (PID-Net) for the yeast counting task with higher accuracy. 

\begin{figure*}
\begin{center}
\includegraphics[width=\textwidth]{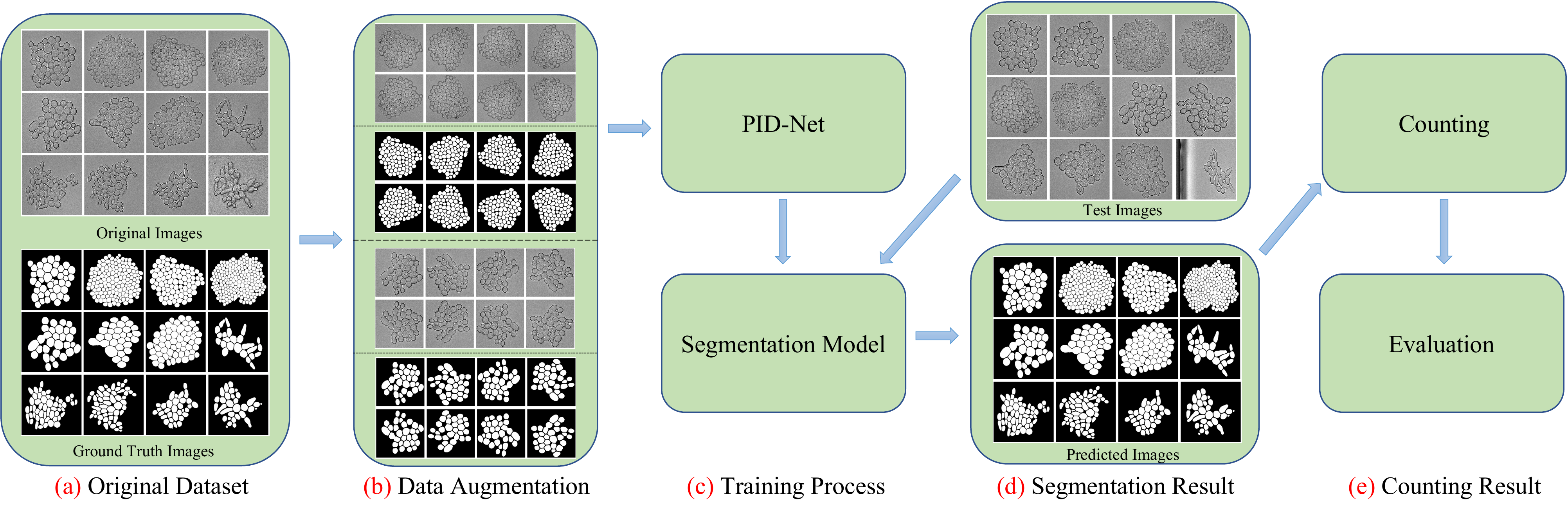}
\caption{The workflow diagram of the proposed yeast image counting method using~PID-Net.}
\label{Fig:workflow}       
\end{center}
\end{figure*}

The PID-Net is an improved Convolutional Neural Network (CNN) based on an encoder--decoder architecture, pixel interval down-sampling and concatenate operations.
By comparing with the traditional SegNet~\cite{Badrinarayanan-2017-SADC} and U-Net~\cite{Ronneberger-2015-UCNB}-based object counting algorithms,  the~accuracy of counting is improved. 
The workflow of the proposed PID-Net counting method is shown in Figure~\ref{Fig:workflow}.

In Figure~\ref{Fig:workflow}, (a) Original Dataset: The dataset contains images of yeast cells and their ground truth (GT). 
The range is from 1 to 256 yeast cells in each image.
(b) Data Augmentation: Mirror and rotation operations are applied to augment the original dataset.
(c) Training Process: PID-Net is trained for image segmentation and the best model is generated. 
(d) Segmentation Result: Test images are processed using the trained PID-Net model and output the predicted segmentation results.
(e) Counting Result: The number of yeast cells is counted by using connected domain~detection.

The main contributions of this paper are as follows:

\begin{itemize}
\item
We propose PID-Net for dense tiny object counting. 
MaxPooling and pixel interval down-sampling are concatenated as down-sampling to extract spatial local and \mbox{global features.} 
%
\item
The operation of max-pooling may lose some local features of tiny objects while segmentation, and~the edge lines may not be connected after max-pooling. 
However, the~PID-Net can cover a more detailed region.
%
\item
The proposed PID-Net achieves better counting performance than other models on the EM (yeast) counting task.
\end{itemize}

The paper is organized as follows: 
Section~\ref{Sec:related-work} is the related work of existing image analysis-based microorganism counting methods. 
Section~\ref{Sec:PID-Net} describes the architecture of the proposed PID-Net in detail. 
Section~\ref{Sec:experiment} consists of the experimental setting, evaluation metrics and results. 
Section~\ref{Sec:conclusion} is the conclusion of this~paper. 

In this section,  related approaches to image analysis-based microorganism counting methods are summarized in Table~\ref{table:relatedworks}, which consist of classical counting methods and machine-learning-based counting methods. 
More detailed research can be referred to in our survey paper~\cite{Li-2021-ACRI}.

\begin{table*}
\caption{Microorganism image counting~methods.}
\label{table:relatedworks}
\begin{tabular}{ccc}
\toprule
\textbf{Category}                 & \textbf{Subcategory}                              & \textbf{Related Work} \\ \midrule
                         & Thresholding-Based Methods               &     \cite{Clarke-2010-LCHT,Zhang-2008-AABC,Zhang-2007-AEAR}         \\
Classical Methods        & Edge Detection-based Methods             &     \cite{Barbedo-2013-AACM,Yamaguchi-2004-MEDC,Ogawa-2003-DMDI}         \\
                         & Watershed-Based Methods                  &        \cite{Ates-2009-AIPB,Selinummi-2005-SQLB,Brugger-2012-ACBC,Masschelein-2012-TACC}      \\ \midrule
                         & Hough Transformation  &  \cite{Austerjost-2017-ASDA,Boukouvalas-2018-ACCS,Alves-2016-CCVA}  \\
Machine-Learning-Methods & Classical Machine-Learning-Based Methods &      \cite{Yoon-2015-ACAC,Zhang-2010-ASTF,Motta-2001-TSPP,Akiba-1997-DASZ}        \\
                         & Deep-Learning-Based Methods              &     \cite{Blackburn-1998-RDBA,Ferrari-2015-BCCC,Ferrari-2017-BCCC,Tamiev-2020-ACBC}       \\  \bottomrule 
\end{tabular}
\end{table*}
\unskip

\subsection{Classical Counting~Methods}

Image segmentation is the most significant part in microorganism-counting task. 
As shown in Table~\ref{table:relatedworks}, the~classical methods contain thresholding, edge detection and watershed methods~\cite{Perez-1987-AITA}. 
For thresholding approaches, the~selection of threshold determines the result of segmentation. 
The most used approaches are iterative thresholding and Otsu thresholding at present. 
Otsu thresholding can achieve satisfactory segmentation results for most images~\cite{Otsu-1979-ATSM}.

Edge detection approaches can extract all boundaries in an image, and~then each close region can be separated~\cite{Magnier-2018-ARSE}. 
Watershed is one kind of region-based segmentation approach, which can be calculated by iterative labeling~\cite{Levner-2007-CDWS}. 
The satisfactory segmentation result can be received though in an image with weak edges.
Hough transformation was proposed for line or circle detection tasks in image with strong anti-noise capability and high accuracy, which can be applied for circular microorgansim counting tasks~\cite{Yuen-1990-CSHT}.

In~\cite{Clarke-2010-LCHT,Zhang-2008-AABC,Zhang-2007-AEAR}, various thresholding methods were applied for microorganism counting. 
In~\cite{Clarke-2010-LCHT}, an~adaptive thresholding was used for microorganism segmentation, and after~that,  the~minima function was applied to locate the center of each colony for counting.
The work~\cite{Zhang-2008-AABC} applied Otsu thresholding for bacteria segmentation, and~the hypothesis testing was then applied for debris erasing.
In~\cite{Zhang-2007-AEAR}, the~contrast-limited adaptive histogram equalization was applied to enhance the plate contour first.
Then, the~Otsu threshold was applied to detect the plate region and binarize the images automatically.  
Finally, the~colonies were separated and~counted.

The works~\cite{Barbedo-2013-AACM,Yamaguchi-2004-MEDC,Ogawa-2003-DMDI} used several edge detection methods for microorganism counting. 
In~\cite{Barbedo-2013-AACM}, five different combination methods applying for microorganism segmentation were compared, such as Gaussian Laplacian and Canny filters. 
Then, the concave surface between the connected colonies was detected for counting. 
The works~\cite{Yamaguchi-2004-MEDC,Ogawa-2003-DMDI} used Sobel and Laplacian filters for the edge detection of bacteria~images.

The works~\cite{Ates-2009-AIPB,Selinummi-2005-SQLB,Brugger-2012-ACBC,Masschelein-2012-TACC} use watershed-based methods for microorganism counting. 
In~\cite{ Ates-2009-AIPB}, watershed was applied for separation of clustered colonies of bacteria images.
After that, the~circularity ratio was calculated for colony counting. 
In~\cite{Selinummi-2005-SQLB}, marker-controlled watershed was applied for bacteria segmentation.
Then, the~number of colonies was estimated as the ratio of cluster area to an average colony area. 
In~\cite{Brugger-2012-ACBC}, Otsu and adaptive thresholding were applied for image binarization.
Then, the~combination method of distance transform and watershed was applied for bacteria counting.
In~\cite{Masschelein-2012-TACC}, watershed was used for image segmentation.
After that, the~gray level co-occurrence matrix (GLCM) of the image was extracted and classified using~SVM.

The works~\cite{Austerjost-2017-ASDA,Boukouvalas-2018-ACCS,Alves-2016-CCVA} used Hough-transformation-based methods for microorganism counting.
In~\cite{Austerjost-2017-ASDA}, the~iterative local threshold was used for bacteria segmentation, and~then a Hough circle transformation was applied to separate clustered colonies into a single colony. 
In~\cite{Boukouvalas-2018-ACCS}, a~median filter was applied for denoising. 
Then, the circular area was detected using the Hough transform to obtain only the inner area of the dish.
Afterward, Gaussian adaptive thresholding was applied for bacteria segmentation.
Finally, cross correlation-based granulometry was used to count the bacteria colonies.
In~\cite{Alves-2016-CCVA}, Otsu thresholding and a Laplacian filter were applied for edge detection.
Then, a~circular Hough transform was used to detect circular bacteria~colonies.

\subsection{Machine-Learning-Based Counting~Methods}
As shown in Table~\ref{table:relatedworks}, the~machine-learning-based microorganism counting approaches consist of machine-learning- and deep-learning-based methods. The~classical machine-learning-based approaches contain Principal Component Analysis (PCA)~\cite{Jolliffe-2016-PCAA} and Support Vector Machine (SVM)~\cite{Vishwanathan-2002-SSVM}. 
Deep-learning methods are usually based on CNN~\cite{Li-2016-SCNN}, Back Propagation Neural Network (BPNN)~\cite{Dai-1997-ELPL} and Artificial Neural Network (ANN)~\cite{Ghate-2010-OMLP} algorithms.

The works~\cite{Yoon-2015-ACAC,Zhang-2010-ASTF,Motta-2001-TSPP,Akiba-1997-DASZ} used classical machine learning for microorganism counting. 
In~\cite{Yoon-2015-ACAC}, PCA was applied for separation the bacteria with other debris.  After~that, the~nearest neighbor searching algorithm was applied for clustered colony separation. 
In~\cite{Zhang-2010-ASTF}, the~shape features were extracted for training, then the SVM was applied for microorganism classification and counting after Otsu thresholding.

In~\cite{Motta-2001-TSPP}, the~histogram local equalization was applied to enhance the contours of protozoa images. 
Then, morphological erosion and reconstruction were used to eliminate the flocs of the protozoa silhouette.  Finally, PCA was applied to classify different protozoa, and~the number of the various species of protozoa was counted. 
In~\cite{Akiba-1997-DASZ}, local auto-correlational masks were used for image enhancement, and PCA was applied for plankton~counting.

The works~\cite{Blackburn-1998-RDBA,Ferrari-2015-BCCC,Ferrari-2017-BCCC,Tamiev-2020-ACBC} used deep learning for microorganism classification and counting. 
In~\cite{Blackburn-1998-RDBA}, the~Marr--Hildreth operator and thresholding were applied for edge detection and binarization of bacteria images. 
Then, ANN was designed for classification and counting. 
In~\cite{Ferrari-2015-BCCC}, the~contrast-limited adaptive histogram equalization was applied for bacteria image segmentation, then four convolutional and one fully connected was trained for classification and counting. 
In~\cite{Ferrari-2017-BCCC}, contrast limited adaptive histogram equalization was used for image enhancement.

Then, CNN was applied for bacteria classification, and~the watershed algorithm was applied for colony separation and counting.
In~\cite{Tamiev-2020-ACBC}, a~classification-type convolutional neural network (cCNN) was proposed for automatic bacteria classification and counting.
First, the~original images were segmented with an adaptive binary thresholding method, and images with individual cells or cell clusters were cropped. 
Then, the images were classified using cCNN. 
The network can output the number of bacteria in given clusters, and~the total count can be~calculated.

By reviewing all related works of microorganism counting, we found that deep learning technologies are widely applied for microorganism classification; however,~few machine-learning-based image segmentation methods are applied. 
Since the development of deep learning and computer vision technologies, CNN-based image segmentation approaches have be applied for accurate microorganism segmentation, such as SegNet~\cite{Badrinarayanan-2017-SADC}, U-Net~\cite{Ronneberger-2015-UCNB}, Attention U-Net~\cite{Oktay-2018-AULW}, Trans U-Net~\cite{Chen-2021-TTMS} and Swin U-Net~\cite{Cao-2021-SUPT}. 
Though the methods above have not been applied for microorganism-counting tasks, they have enough potential to extract the microorganism before counting, which can be inferred with better~performance.

Since the yeast images in our dataset  range from 1 to 256 yeast cells in each image, and~the boundaries of the cells are not clear, it can be inferred that the classical segmentation methods may show poor performance in this counting task. 
Therefore, we propose an encoder--decoder model that concentrates on the dense and tiny object counting~task.

\section{PID-Net-Based Yeast Counting~Method}\label{Sec:PID-Net}
\label{deeplearningmethods}

Although the existing image segmentation models, such as SegNet and U-Net, have been widely applied in semantic segmentation and biomedical image segmentation, they still cannot meet the requirements of accurate segmentation in the microorganism-counting task. 
To this end, we propose PID-Net, a~CNN-based on pixel interval down-sampling, MaxPooling and concatenate operations to obtain a better performance. 
The process of microorganism counting mainly contains two parts, the~first  is microorganism image segmentation, whose purpose is to classify the foreground and background at the pixel-level. 
The second part is microorganism counting, whose purpose is to count the number of segmented objects after~post-processing.

\subsection{Basic Knowledge of~SegNet}
SegNet is a CNN-based image segmentation network with the structure of an encoder and decoder. 
The innovation of SegNet is that the~dense feature maps of high resolution images can be calculated by the encoder, and~the up-sampling operation for low-resolution feature maps can be performed by the decoder network~\cite{Badrinarayanan-2017-SADC}. 
The structure of SegNet can be considered as an encoder network and a corresponding decoder network. The~last part is a pixel-level classification layer. 

The first 13 convolutional layers of VGG16~\cite{Simonyan-2014-VDCN} is applied in encoder network of SegNet, which consist of convolutional layers, pooling layers and Batch Normalization layers. 
In the encoder network, two sequences, which consist of one 3 $\times$ 3 convolution operation, followed by a Batch Normalization and a ReLU operation, are applied in each step. 
After that, the~feature maps are down-sampled by using a max-pooling operation with the size of 2 $\times$ 2 and stride of 2 pixels. 
After pooling, the~size of the feature map is changed into half of the initial.

What is noteworthy is that the Pooling Indices are saved while pooling, which records the initial position of the maximum value in the input feature maps.
In the decoder network, the up-sampling operation is applied for feature maps, and then the convolution operation is performed three times to fix the detail loss while pooling. 
The same operation is replicated five times to change the feature maps into the initial image size. 
The saved Pooling Indices are applied while up-sampling to set the feature points into correct positions. 
A Softmax layer is applied finally for feature map~classification.

\subsection{Basic Knowledge of~U-Net}

U-Net is an U-shape CNN model based on an encoder--decoder and skip connection. 
U-Net is first designed for the segmentation of biomedical images. 
The max-pooling with the size of 2 $\times$ 2 and stride of 2 pixels is applied for down-sampling. 
There are two 3 $\times$ 3 convolution operations (each followed by a ReLU) between two down-sampling operations. 
The down-sampling operation repeats four times, and the number of feature map channels is modified to 1024. 

In the decoder network, the~result after up-convolution operation (a 2 $\times$ 2 up-sampling and a 2 $\times$ 2 convolution operation) is concatenated with the corresponding feature maps of encoder, which can combine the high-level semantics with the low-level fine-grained information of the image. 
After that, two 3 $\times$ 3 convolution operations (each followed by a ReLU) are applied. 
The size of each feature map is changed into the size of input after four up-sampling operations. 
Finally, a~Sigmoid function is applied for~classification.

\subsection{The Structure of~PID-Net}
\begin{figure*}
\centering
\includegraphics[trim={0cm 0cm 0cm 0cm},clip,width=1.0\textwidth]{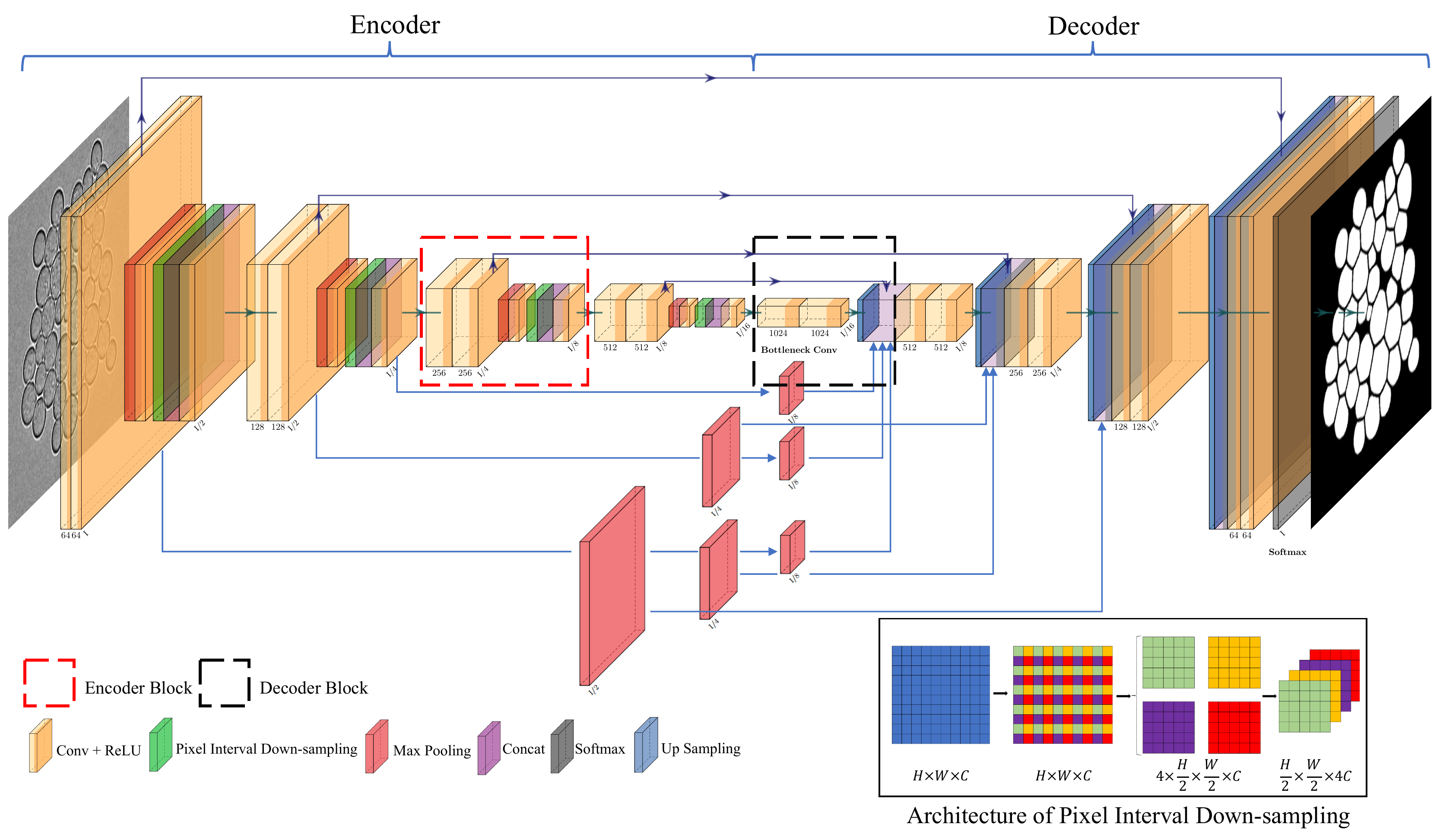}
\caption{The structure of the proposed~PID-Net.}
\label{Fig:sscnet}
\end{figure*}

Following the basic idea of SegNet and U-Net for image segmentation, the~structure of the proposed PID-Net is shown in Figure~\ref{Fig:sscnet}, which is an end-to-end CNN structure based on the encoder and decoder.
There are four blocks in the encoder network. 

The first parts in each block are two convolution operations with a kernel size of 3 $\times$ 3 (each followed by a ReLU operation), and then the max-pooling with the size of 2 $\times$ 2 and stride of 2 pixels is applied to reduce the size of feature maps by half, followed by a convolution and ReLU operation.
The channel of feature maps is denoted as $C$.
Pixel interval down-sampling is applied for down-sampling, which is shown in Figure~\ref{Fig:sscnet}. 
Each pixel is sampled with the pixels apart, and the size of each feature map is replaced by half.

The classical down-sampling methods, such as max-pooling (with the kernel size of 2) will drop $\frac{3}{4}$ data of the original image. 
It can retain the main information but not be fit to the task of tiny object counting (the edge lines may be lost while max-pooling). 
Though there are several learnable pooling layers that have proposed, such as Fractional pooling~\cite{Graham-2014-FMPO}, Stochastic pooling~\cite{Zeiler-2013-SPRD} and~learned-norm pooling approaches~\cite{Gulcehre-2014-LPDF}; however,~they still cannot meet the requirement of accurate segmentation for dense tiny yeast cells.

Thus, a new down-sampling method, pixel interval down-sampling, is proposed here, which can reduce the size of feature maps without dropping data.
Afterward, four pixel interval down-sampling feature maps and the features after max-pooling are concatenated to 5$C$-dimensional features. 
Finally, a~convolutional filter with $C$ channels is applied to reduce  5$C$-dimensional features to $C$-dimensional features. 
Hereto, the~initial feature maps with size H $\times$ W and channel $C$ are changed to feature maps with size $\frac{H}{2}$ $\times$ $\frac{W}{2}$ and channel $C$. 
The procedure is repeated four times  with~output resolutions of $\frac{H}{16}$ $\times$ $\frac{W}{16}$ and channel of 8$C$.

In the decoder network, four blocks are applied for up-sampling. 
Two convolution operations with  a kernel size of 3 $\times$ 3 (each followed by a ReLU operation) are applied first. 
Then, the transposed convolution operation with a kernel size of 3, a~stride of 2 and~padding of 1 is applied for up-sampling. 
The transposed convolution operation is widely applied in GANs to expand the size of images~\cite{Zeiler-2011-ADNM}. 
The count of channels after up-sampling of the bottleneck is 512, which is calculated by using the transposed convolutional filter with 512 channels~\cite{Zeiler-2010-DENE}.

The params in the transposed convolution filter can be learned while training.
After that, the~high resolution feature maps of  encoder network are transformed to low-resolution feature maps using 2 $\times$, 4 $\times$ and 8 $\times$ max-pooling, which is shown in Figure~\ref{Fig:sscnet}.
Then, the~feature maps after up-sampling and max-pooling are concatenated with the feature maps generated by the corresponding layer from the encoder. 

For instance, the~8 $\times$ max-pooling features of the first block, 4 $\times$ max-pooling of the second block and 2 $\times$ max-pooling of the third block in the encoder are concatenated with the copied features of the fourth encoder block and the features after up-sampling (five parts of feature maps are concatenated in the first decoder block). 
In the same way, there are 4, 3 and 2 parts of features are concatenated in the second, third and~fourth level of the decoder, respectively.
After the concatenated operation, two convolutions and ReLU operations are applied to change the number of channels. 
The up-sampling operation is repeated four times with~output resolutions of $H$ $\times$ $W$ and channel of $C$, which has the same size as the encoder’s input features.
Finally, a~Softmax layer with two output channels is applied for feature map~classification.

\subsection{Counting~Approach}

\begin{figure*}
\centering
\includegraphics[trim={0cm 0cm 0cm 0cm},clip,width=1.0\textwidth]{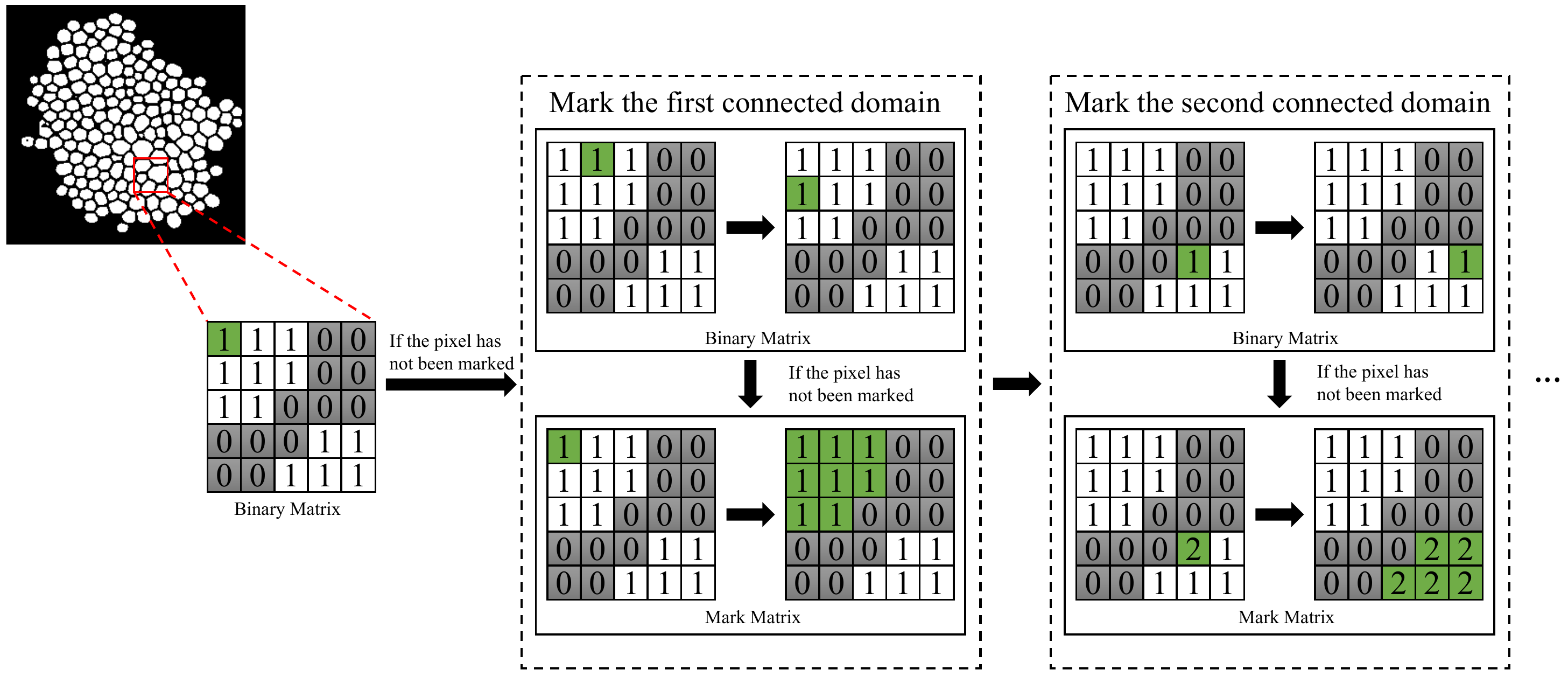}
\caption{An example of the counting process based on the eight neighborhood~search.}
\label{Fig:counting}
\end{figure*}

A post-processing method is applied to eliminate the effect of noises after segmentation. 
First,  a~morphological filter is applied to remove useless debris, which can improve the performance of counting prominently. 
Then, the eight neighborhood search algorithm is applied to count the connected regions of segmented images after denoising~\cite{Boss-2013-AMIB}.
The process of counting is shown in Figure~\ref{Fig:counting}. 
A binary matrix is traversed in line and a mark matrix is applied to mark the connected domain~\cite{Wang-2006-NNNC}.
Finally, the~number of connected domain in mark matrix is the number of yeast~cells.

\section{Experiments}\label{Sec:experiment}

\subsection{Experimental~Setting}
\label{expersetting}

\subsubsection{Image~Dataset}\label{4.1.4}
\begin{figure*}
\centering
\includegraphics[trim={0cm 0cm 0cm 0cm},clip,width=1.0\textwidth]{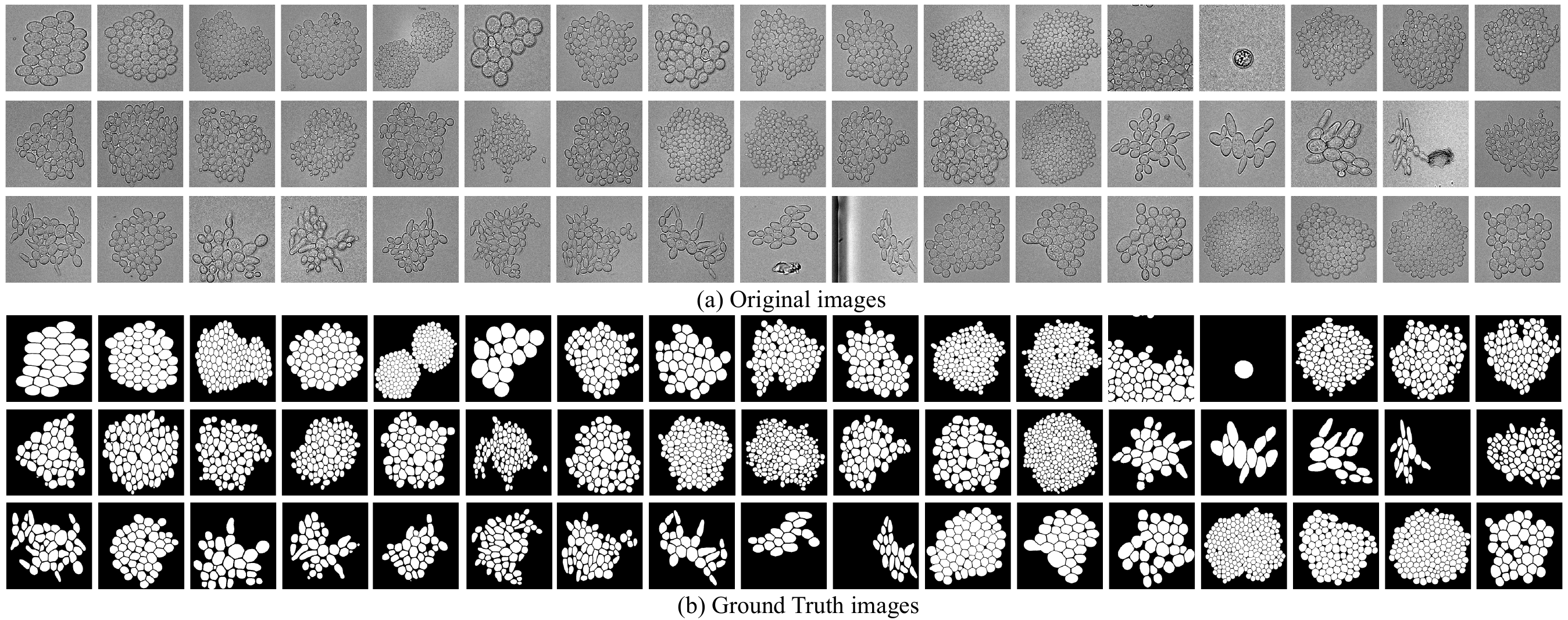}
\caption{The images in yeast cell dataset. (\textbf{a}) The original yeast image and (\textbf{b})  the corresponding ground truth~images.}
\label{Fig:gtimg}
\end{figure*}

In our work, we use a yeast image dataset proposed in~\cite{Dietler-2020-ACNN}, containing 306 different images of yeast cells and their corresponding ground truth (GT) images. 
All images are resized to the resolution of 256 $\times$ 256 pixels, which are shown in Figure~\ref{Fig:gtimg}.
Then, the original 306 images are rotated (0, 90, 180 and 270 degrees) and flipped (mirror), and thus the number of images in this dataset is augmented to eight times (2448 images). 

\unskip

\subsubsection{Training, Validation and Test Data~Setting}

The original yeast image dataset was randomly divided into training, validation and test dataset with the ratio of 3:1:1, and then, each dataset was augmented  eight times.
Therefore, there  1470 images with their GT were applied as the training dataset, 489 images with their corresponding GT were applied for validation, and~489 original images were applied for testing.

\subsubsection{Experimental~Environment}
The experiment was conducted by Python 3.8.10 in Windows 10 operating system. 
The experimental environment was based on Torch 1.9.0. 
The workstation was equipped with Intel(R) Core(TM) i7-8700 CPU with 3.20GHz, 16GB RAM and~NVIDIA GEFORCE RTX 2080~8GB.

\subsubsection{Hyper~Parameters}
In the experiment of yeast cell counting, the~purpose of image segmentation is to determine whether a pixel is a foreground (yeast cell) or background. 
The last part in the proposed PID-Net before the output is Softmax, which is applied to calculate the classification result of feature maps. 
The definition of Softmax is shown as Equation~(\ref{eq:softmax}).
\begin{equation}
{\rm Softmax}(z_{i}) =\frac{e^{z_{i}}}{\sum_{c=1}^{C}e^{z_{c}}} .
\label{eq:softmax}
\end{equation}

In Equation~(\ref{eq:softmax}), $z_{i}$ is the output in the $i$th node, and $C$ is the number of output nodes, representing the number of classified categories. 
The classification prediction can be converted into the probabilities by using the Softmax function, which distributes in the range of [0, 1], and~the sum of probability is 1.
As the image segmentation for yeast counting is to distinguish the foreground and the background. Hence, it is a binary classification, Equation (\ref{eq:softmax}) can be rewritten as Equation~(\ref{eq:sigmoid}).
\begin{equation}
{\rm Softmax}(z_{1})=\frac{e^{z_{1}}}{e^{z_{1}}+e^{z_{2}}}=\frac{1}{1+e^{-(z_{1}-z_{2})}}={\rm Sigmoid}(\beta) .
\label{eq:sigmoid}
\end{equation}

In Equation~(\ref{eq:sigmoid}), $\beta$ is $(z_{1}-z_{2})$, which means the Softmax function and the Sigmoid function are the same for binary classification (a little difference between them is, the~number of the fully connected (FC) layer of Softmax is two to distinguish two different categories; however,~the number of FC layer of Sigmoid is one, only to judge whether the single pixel is the object to be segmented).

The probability of the pixel to be classified as 1 is:
\begin{equation}
\hat{y}=P(y=1|x) .
\end{equation}

Apparently, 
the~probability of the pixel to be classified as 0 is:
\begin{equation}
1-\hat{y}=P(y=0|x) .
\end{equation}

According to the maximum likelihood formula, the~joint probability can be expressed as:
\begin{equation}
P(y|x)=\hat{y}^{y}\cdot(1-\hat{y})^{1-y} .
\end{equation}

After that, $log$ function is applied to remain the monotonicity invariance of the function:
\begin{equation}
logP(y|x)=log(\hat{y}^{y}\cdot(1-\hat{y})^{1-y})=ylog\hat{y}+(1-y)log(1-\hat{y}) .
\end{equation}

Hereto, the~loss can be expressed as $-logP(y|x)$, and~the loss function for multiple samples can be defined as the cross-entropy loss ($N$ is the number of categories):
\begin{equation}
Loss=-[\sum_{i=1}^{N}y^{(i)}log\hat{y}^{(i)}+(1-y^{(i)})log(1-\hat{y}^{(i)})] .
\end{equation}

In order to guarantee the stable and fast convergence of the proposed network, we deploy preliminary experiments to determine the choices of hyper parameters.
Adaptive moment estimation (Adam) is compared with stochastic gradient descent (SGD) and natural gradient descent (NGD).
Adam optimizer has the smoothest loss curves and stablest convergence, which performs best in microorganism-counting task.
Adam optimizer is applied to minimize the loss function, which can adjust the learning rate automatically by considering the gradient momentum of the previous time steps~\cite{Kingma-2014-AAMS}.

\begin{figure*}
\centering
\includegraphics[trim={0cm 0cm 0cm 0cm},clip,width=1.0\textwidth]{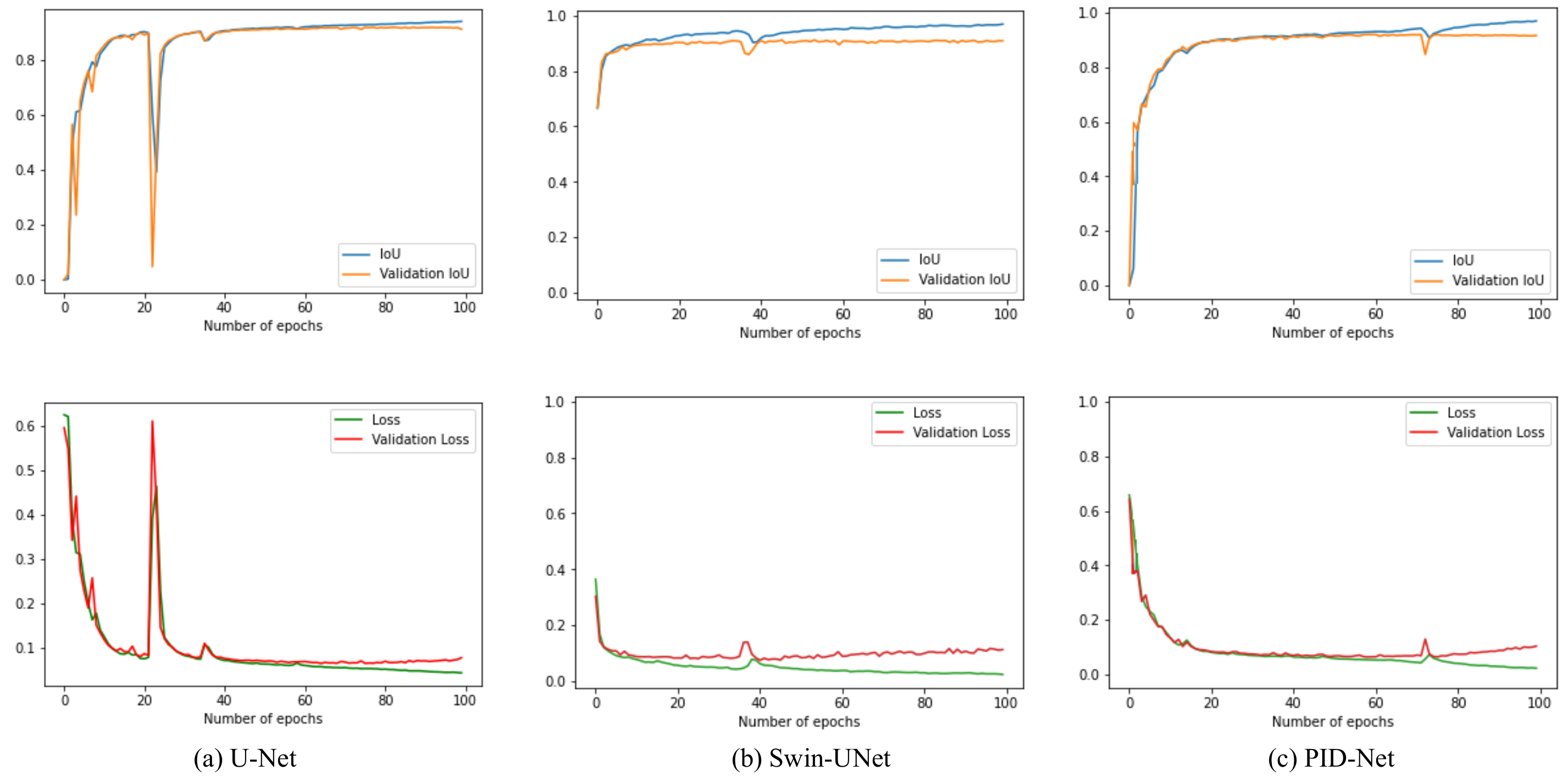}
\caption{The IoU (\textbf{top row}) and loss (\textbf{bottom row}) curves in the training~process.}
\label{Fig:lossimg}
\end{figure*}

The initial learning rate is set from 0.0001 to 0.01 in preliminary experiments. 
By observing the loss curves while training, the~learning rate of 0.001 can balance the speed and stability of convergence.
The batch size is set as 8 due to the limited memory size (8GB).
The selection of the hyper parameters above are optimal in preliminary experiments, and thus they are applied in our formal microorganism counting experiment.
The epoch is set as 100 by considering the converge speed of experimental models, the~example of loss and intersection over union (IoU) curves of models is shown in Figure~\ref{Fig:lossimg}. 

Though there are 92,319,298 params to be trained in PID-Net; however,~it can converge rapidly and smoothly without~over fitting.
There is a jump in loss and IoU plots for all three tested networks from 20 to 80 epochs, which is caused by the small batch size. Small batch size may lead to huge difference between each batch, and~the loss and IoU curves may jump with~convergence.

\unskip

\subsection{Evaluation~Metrics}

In the task of dense tiny object counting, the~evaluation of image segmentation is the most significant part. 
Hence, the~widely applied segmentation evaluation metrics Accuracy, Dice, Jaccard and Precision are employed here to evaluate the performance of microorganism segmentation. 
Furthermore, the Hausdorff distance is applied to evaluate the shape similarity between the predicted image and GT. 
Finally, the~counting accuracy is calculated to quantify the counting performance of the~models.

Accuracy is applied to calculate the proportion of pixels that are correctly classified. 
The Dice coefficient~\cite{Dice-1945-MTAE} is applied to measure the similarity of the predicted image and GT. The~similarity can be quantified range from 0 to 1 (1 means the predicted result coincides exactly with the GT).
Jaccard~\cite{Jaccard-1912-TDTF}, also named the intersection over union (IoU), is applied to compare the similarity and differences between the predicted image and GT image, focusing on whether the samples’ common characteristics are consistent.
Precision is defined as the proportion of positive pixels in the pixels, which are classified as positive. 

\begin{figure}[H]
\includegraphics[trim={0cm 0cm 0cm 0cm},clip,width=0.5\textwidth]{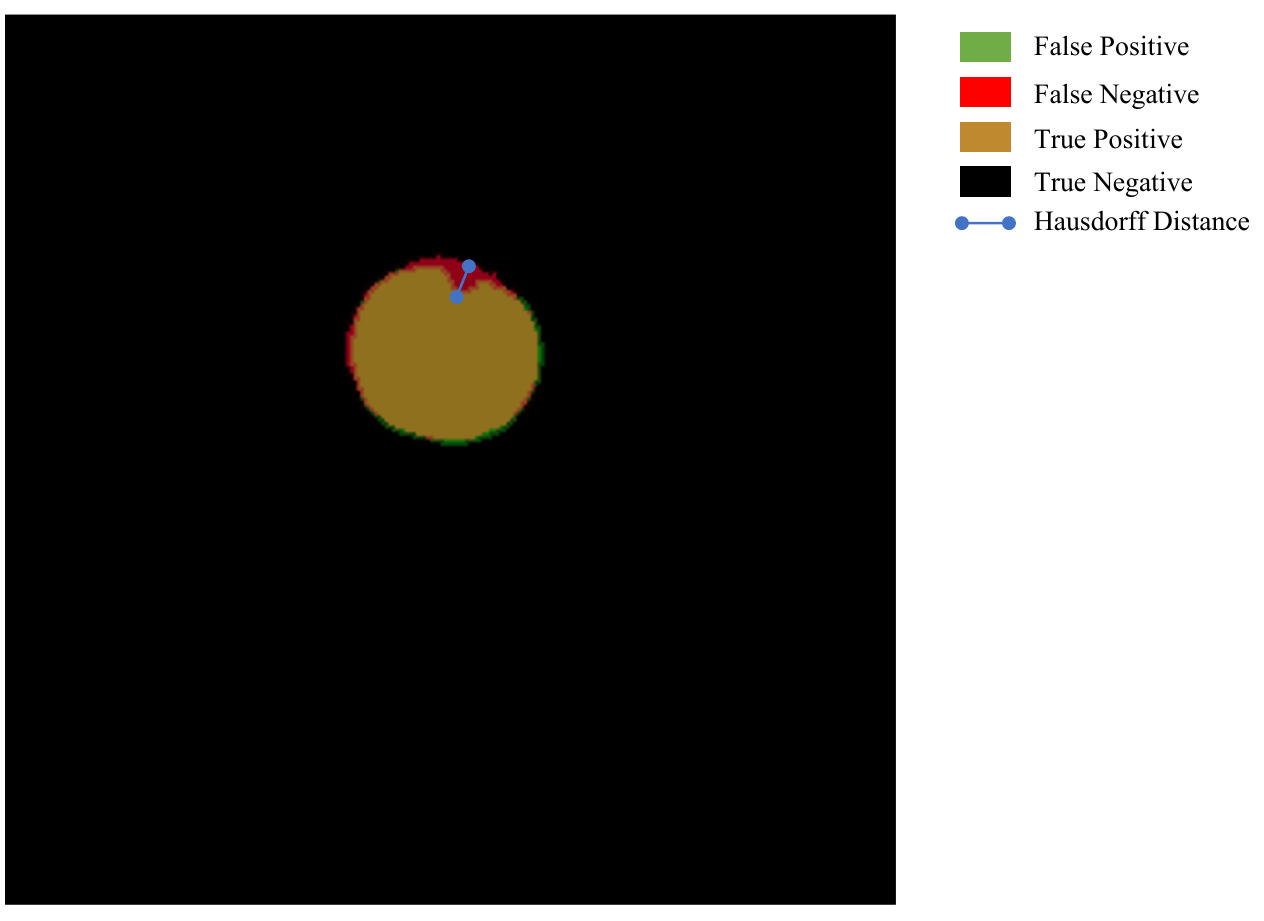}
\caption{The visualization result of the Hausdorff distance between the GT and predicted~image.}
\label{Fig:hausdorff}
\end{figure}

The Hausdorff distance~\cite{Huttenlocher-1993-CITH} is applied to measure the Euclidean distance between the predicted and GT images with~the unit of pixels in per image.
In contrast with Dice, the Hausdorff distance focuses on the boundary distance of two objects to measure the shape similarity; however,~the Dice majors in the inner similarity. 
An example of the Hausdorff distance between GT and predicted image is shown in Figure~\ref{Fig:hausdorff}.
The Hausdorff is the maximum of the shortest distance between a pixel in a image and another image~\cite{Sim-1999-OMAR}.

In the task of microorganism counting, the Hausdorff distance can be applied to measure the shape similarity between the GT and segmentation result, showing the performance of segmentation models.
Finally, the~performance of counting is measured using counting accuracy, which is defined as the proportion of the predicted number and GT number of yeast cell~images.

\begin{figure*}
\centering
\includegraphics[trim={0cm 0cm 0cm 0cm},clip,width=1.0\textwidth]{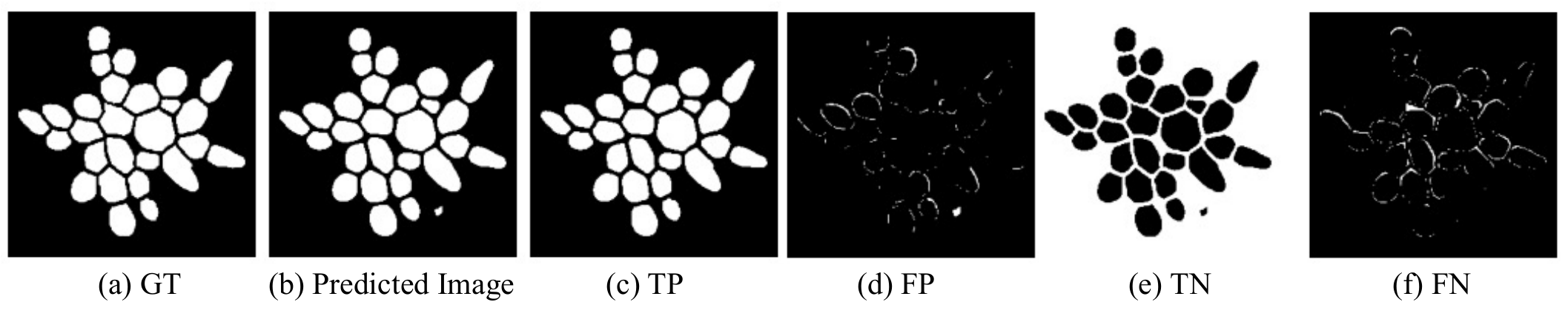}
\caption{The illustration of TP, TN, FP and FN between the predicted image and GT~image.}
\label{Fig:ttpp}
\end{figure*}

The definitions of the proposed evaluation metrics are summarized in Table~\ref{table:metrics}. 
The TP (True Positive), TN (True Negative), FP (False Positive) and FN (False Negative) are basic evaluation metrics, which can be applied to measure the performance of segmentation in general. 
An example of a yeast cell image with its TP, TN, FP and FN is illustrated in Figure~\ref{Fig:ttpp} for intuitive understanding.
$V_{pred}$ is the foreground after segmentation by using the model, $V_{GT}$ is the foreground of the GT image. 
Furthermore, $N_{pred}$ means the number of connected regions in the predicted image, $N_{GT}$ means the number of connected regions in the GT image, which indicates the number of yeast cells. 
In the definition of the Hausdorff distance, $sup$ is the supremum, and $inf$ is the~infimum.

\begin{table*}
\footnotesize\setlength{\tabcolsep}{2.91mm}
\caption{The definitions of evaluation metrics. CA and HD are abbreviations of the Counting Accuracy and Hausdorff Distance, respectively. }
\label{table:metrics}
\begin{tabular}{cccc}
\toprule
\textbf{Metric}            & \textbf{Definition}  & \textbf{Metric}            & \textbf{Definition} \\ \midrule 
Accuracy          &   $\frac{\rm TP+TN}{\rm TP+TN+FP+FN}$   & Dice    &    $\frac{2\times |V_{pred}\bigcap V_{GT}|}{|V_{pred}| + |V_{GT}|}$       \\  \midrule 
Jaccard           &  $\frac{|V_{pred}\bigcap V_{GT}|}{|V_{pred}\bigcup V_{GT}|}$ & Precision & $\frac{\rm TP}{\rm TP+FP}$ \\  \midrule 
CA & $1-\frac{|N_{pred}-N_{GT}|}{N_{GT}}$  & HD & $d_H(X,Y)=max(sup_{x\in X}inf_{y\in Y}d(x,y), sup_{y\in Y}inf_{x\in X}d(x,y))$ \\ 
\bottomrule
\end{tabular}
\end{table*}

The proposed evaluation metrics, containing the Accuracy, Dice, Jaccard and Precision, are proportional to the segmentation performance of models.  
The Hausdorff distance has an inverse correlation with the  segmentation performance.
Counting accuracy can evaluate the final counting results of different~models. 

\subsection{Evaluation of Segmentation and Counting~Performance}

To prove the satisfactory segmentation performance of the proposed PID-Net for dense tiny object counting, we compare different down-sampling methods to show the advancement of our proposed method.
Furthermore, several state-of-the-art approaches are applied for comparative experiments. 
All of the experimental setting and evaluation indices are same for comparative experiment. Furthermore, the same dataset is applied for all comparative experiments, which is proposed in Section~\ref{4.1.4}. 
The models are trained from scratch without pre-training and~fine-tuning. 

\subsubsection{Comparison of Different Down-Sampling~Methods}

In this part, we compare the effect of different down-sampling and skip connection approaches for segmentation. 
In our proposed PID-Net, pixel interval down-sampling and max-pooling operations are concatenated to combine the dense and sparse feature maps after convolution operations.
Then, in the process of hierarchy skip connection, max-pooling is applied to combine the high-level features and low-level features directly, which is beneficial to reduce the effect of resolution loss while up-sampling and help rebuild the segmentation result.

\begin{figure}[H]
\includegraphics[trim={0cm 0cm 0cm 0cm},clip,width=0.5\textwidth]{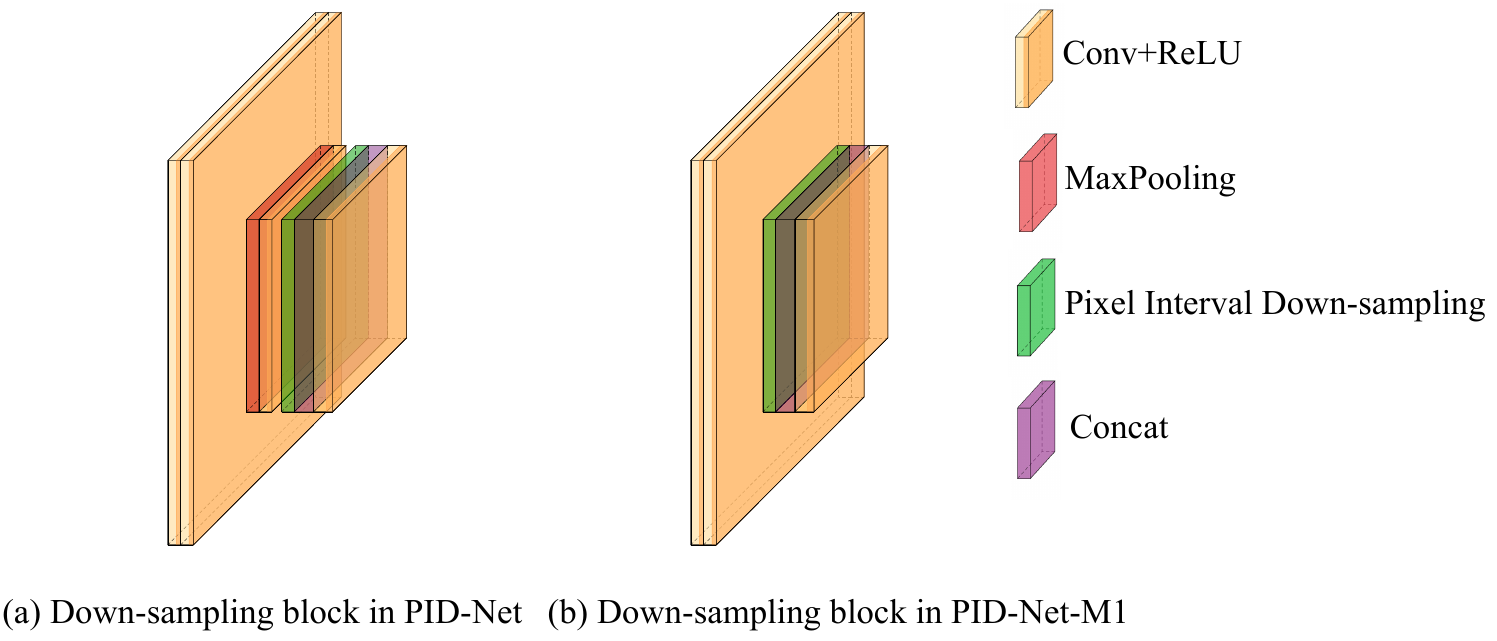}
\caption{The down-sampling block of PID-Net and~PID-Net-M1.}
\label{Fig:sscM1}
\end{figure}

To show the effectiveness and reasonability of the proposed method, we change the approaches of down-sampling and hierarchy skip connection as  PID-Net Modified-1 (PID-Net-M1) and PID-Net Modified-2 (PID-Net-M2).
In PID-Net-M1, max-pooling operations are only applied in the process of hierarchy skip connection and~not in down-sampling. 
The down-sampling block of PID-Net-M1 is illustrated in Figure~\ref{Fig:sscM1}.
In PID-Net-M2, all down-sampling operations are realized using pixel interval down-sampling without max-pooling.
The segmentation evaluations and counting performance of those approaches are shown in Table~\ref{table:compare}.

From Table~\ref{table:compare}, we find that the proposed PID-Net achieves the best counting performance. 
By comparing with the PID-Net-M1 and PID-Net-M2, the~average accuracy is increased by 0.1\% to 0.6\%; 
the improvement of average Dice value is 0.1\% to 1.1\%; 
the average Jaccard is improved by around 0.3\% to 1.6\%. 
Furthermore, the~mean Hausdorff distance of PID-Net is the shortest, which indicates the similarity between the predicted images and GT images is the highest. 
Finally, the~counting accuracy achieved  96.97\%, which shows the satisfactory counting performance of the PID-Net. 
Hence, the~segmentation and counting performance of PID-Net is the best by referring to all evaluation~metrics. 
\unskip

\begin{table*}
\caption{The average segmentation evaluation indices of predicted images. A, D, J, P, C and H are abbreviations of the Accuracy, Dice, Jaccard, Precision, Counting Accuracy (in \%) and Hausdorff Distance (in pixels/per image), respectively. }
\label{table:compare}

\begin{tabular}{ccccccc}
\toprule
\textbf{Methods}   & \textbf{A}     & \textbf{D}     & \textbf{J}     & \textbf{P}     & \textbf{C}       & \textbf{H}      \\ 
\midrule
PID-Net                  & 97.51 
& 95.86 & 92.10 & 96.02 & 96.97 & 4.6272 \\ 
PID-Net-M1               & 96.90 & 94.71 & 90.05 & 94.71 & 67.84   & 5.0110 \\ 
PID-Net-M2               & 97.42 & 95.75 & 91.89 & 95.68 & 96.88   & 4.7204 \\
\bottomrule
\end{tabular}
\end{table*}

\subsubsection{Comparison with Other~Methods}

In this part, some comparative experiments are applied for the yeast cell counting task. 
Some classical methods proposed in Section~\ref{classicalmethods} and deep-learning-based methods proposed in Section~\ref{deeplearningmethods} are compared, consisting Hough transformation~\cite{Illingworth-1988-ASTH}, Otsu thresholding, Watershed, SegNet and U-Net-based segmentation approaches. 
Furthermore, we conduct some extra experiments using  state-of-the-art approaches, containing Attention U-Net~\cite{Oktay-2018-AULW}, Trans U-Net~\cite{Chen-2021-TTMS} and Swin U-Net~\cite{Cao-2021-SUPT}.

\begin{figure*}
\centering
\includegraphics[trim={0cm 0cm 0cm 0cm},clip,width=1.0\textwidth]{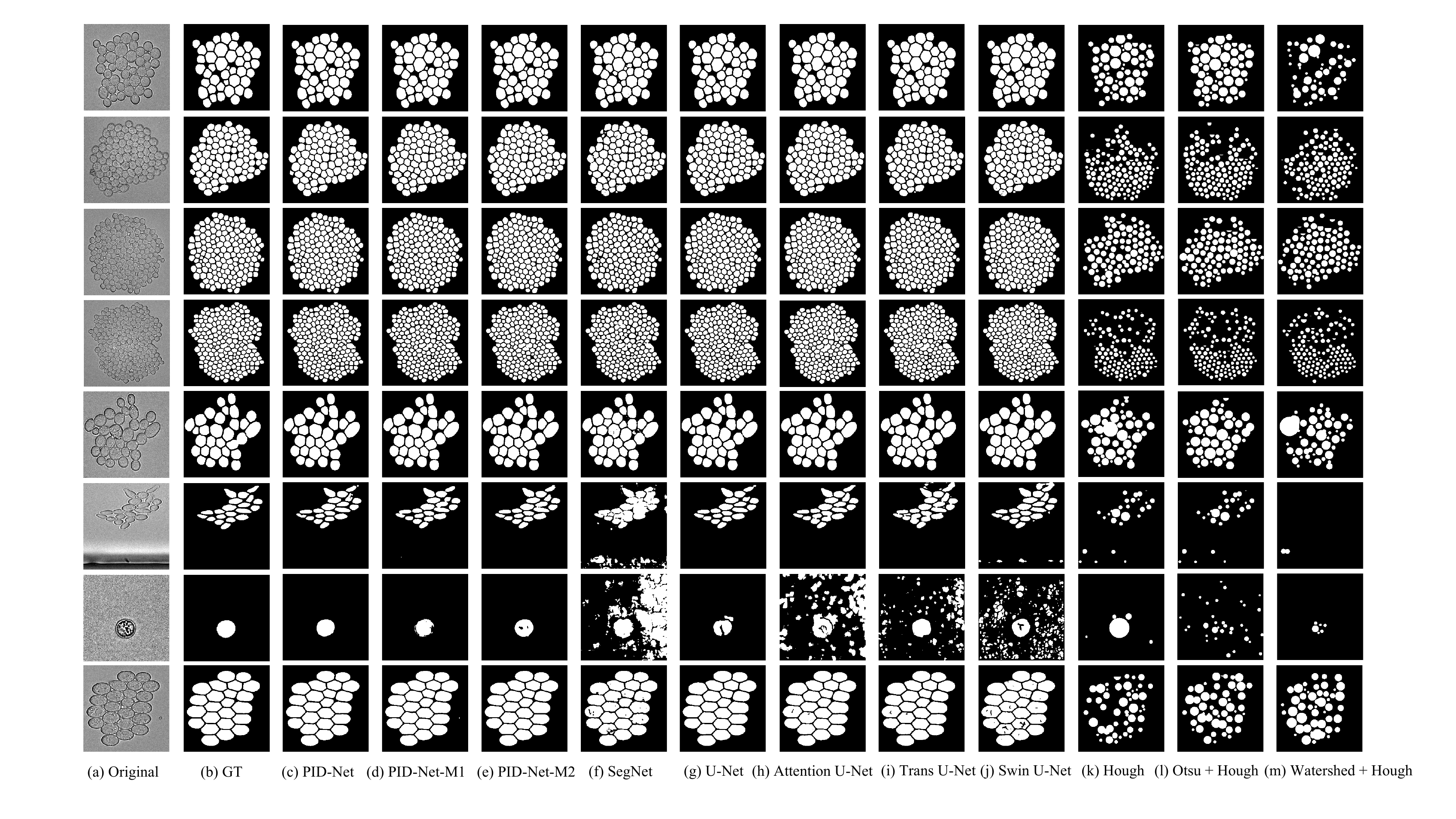}
\caption{An example of segmentation images predicted by different~models.}
\label{Fig:modelscompare}
\end{figure*}

Due to the determination of $k$ in clustering methods, such as $k$-means, is still an insoluble problem while counting; therefore, the clustering-based approaches cannot be applied here for dense tiny object counting. 
All comparative experiments have the same experimental setting, which can be referred to Section~\ref{expersetting} for details.
After image segmentation and object counting, the~average evaluation indices are summarized in Table~\ref{table:finalresult}, and the example images of segmentation are shown in Figure~\ref{Fig:modelscompare}.

\begin{table*}
\caption{The average segmentation evaluation indices of predicted images. A, D, J, P, C and H are abbreviations of the Accuracy, Dice, Jaccard, Precision, Counting Accuracy (in \%) and Hausdorff Distance (in pixels/per image), respectively. }
\label{table:finalresult}
\begin{tabular}{ccccccc}
\toprule
\textbf{Methods}   & \textbf{A}     & \textbf{D}     & \textbf{J}     & \textbf{P}     & \textbf{C}       & \textbf{H}      \\ 
\midrule
PID-Net                  & 97.51 & 95.86 & 92.10 & 96.02 & 96.97 & 4.6272 \\ 
SegNet                   & 94.69 & 90.34 & 84.02 & 88.50 & 68.82 & 6.3604 \\
YeaZ (in~\cite{Dietler-2020-ACNN}) & -& 94.00 & - & - & - & - \\
U-Net                    & 97.47 & 95.71 & 91.84 & 95.62 & 91.33   & 4.6666 \\
Attention U-Net           & 96.62 & 93.36 & 88.96 & 92.67 & 83.44 & 5.1184 \\
Trans U-Net        & 96.84 & 93.60 & 88.99 & 93.25 & 91.32 & 5.0715 \\
Swin U-Net                & 96.47 & 92.99 & 88.32 & 92.43 & 91.95 & 5.3140 \\
Hough          & 82.12 & 61.12 & 44.74 & 88.26 & 73.66   & 9.2486 \\
Otsu                     & 84.23 & 65.71 & 49.90 & 87.66 & 74.34    & 8.9165 \\
Watershed                & 78.67 & 50.15 & 34.88 & 78.61 & 63.34   & 9.6873 \\

\bottomrule
\end{tabular}
\end{table*}

From the evaluation indices summarized in Table~\ref{table:finalresult}, we can find that the PID-Net has the highest Accuracy, Dice, Jaccard, Precision and Counting Accuracy and~the lowest Hausdorff distance, which means the proposed model performs best in the task of dense tiny object counting by comparing with other models. 
Even more, the~Jaccard of PID-Net is higher than the YeaZ who proposed this yeast cell dataset.
In general, the~approaches based on deep learning perform better than the classical approaches.

We  find that the Counting Accuracy of several methods are very low abnormally, which may caused by the enormous difference between the GT and the predicted image.
For instance, the~single yeast cell image in Figure~\ref{Fig:modelscompare} performs unsatisfactory when the segmentation is not accurate. 
The segmentation results of SegNet and Attention-UNet have a large number of False Positive pixels, and~the counting approach is based on the connected domain detection; hence, the value of $\frac{|N_{pred}-N_{GT}|}{N_{GT}}$ is much  higher than~normal.

From the best performance of the proposed PID-Net in the task of dense tiny object counting, we can infer that the down-sampling and skip connection part of PID-Net, which combines max-pooling and pixel interval down-sampling can obtain the feature maps of dense tiny objects and reconstruct the images~better.

\subsection{Repeatability~Tests}

Five additional experiments were repeated based on the original PID-Net model for repeatability tests.
The evaluation indices are given in Table~\ref{table:repeatresult}. 
From Table~\ref{table:repeatresult}, we find that all evaluation indices of repeated PID-Nets are approximate, which shows satisfactory and stable counting performance for the dense tiny object counting~task. 

\begin{table*}
\caption{The evaluation indices of Repeatability Tests. A, D, J, P, C and H are abbreviations of the Accuracy, Dice, Jaccard, Precision, Counting Accuracy (in \%) and Hausdorff Distance (in pixels/per image), respectively. }
\label{table:repeatresult}
\begin{tabular}{ccccccc}
\toprule
\textbf{Methods}   & \textbf{A}     &\textbf{ D}     &\textbf{ J}     & \textbf{P}     & \textbf{C}       & \textbf{H}      \\ \midrule
PID-Net                  & 97.51 & 95.86 & 92.10 &96.02 & 96.97 & 4.6272 \\
PID-Net (Re 1)       & 97.51 & 95.79 & 91.97 & 95.91 & 95.26   & 4.5865 \\
PID-Net (Re 2)      & 97.33 & 95.59 & 91.62 & 95.70 & 96.25   & 4.7290 \\
PID-Net (Re 3)      & 97.54 & 95.91 & 92.18 & 96.21 & 96.82 & 4.6023 \\
PID-Net (Re 4)      & 97.37 & 95.64 & 91.70 & 95.70 & 95.51   & 4.7471 \\
PID-Net (Re 5)      & 97.43 & 95.66 & 91.73 & 92.24 & 96.26 & 4.6395 \\
 \bottomrule
\end{tabular}
\end{table*}
\unskip

\subsection{Computational~Time}

The training time, mean training time, test time and~mean test time are listed in Table~\ref{table:traintime}. 
There are 1470 images in the training dataset and 489 images in the test dataset. 
The mean training time of PID-Net model is approximately 2.9 seconds higher than the time of U-Net, and~the test time is about 0.4 seconds higher than U-Net. 
The memory cost of PID-Net is about 20MB, which is about 6 MB more than the cost of U-Net model, meanwhile, the~PID-Net has better counting performance and lower memory cost than Swin-UNet (41~MB).
The counting accuracy is increased about 6\%; hence, the PID-Net has  satisfactory counting performance and a tolerable computational time, which can be widely applied in accurate dense tiny object counting~tasks.

\begin{table*}
\caption{The summary of computational time (in seconds).}
\label{table:traintime}
\begin{tabular}{ccccc}
\toprule
\textbf{Model}          & \textbf{Training Time} & \textbf{Mean Training Time}& \textbf{Test Time}& \textbf{Mean Test Time} \\ \midrule
PID-Net        & 10,438.86       & 7.10       &     454.68      &        0.93           \\
U-Net          & 6198.00       & 4.21        &     257.64      &          0.53         \\
Swin-UNet      & 7884.36       & 5.36      &       319.50    &          0.65         \\
Att-UNet & 6983.58       & 4.75    &     296.64      &       0.61            \\ \bottomrule
\end{tabular}
\end{table*}
\unskip

\subsection{Discussion}

\begin{figure*}
\centering
\includegraphics[trim={0cm 0cm 0cm 0cm},clip,width=1.0\textwidth]{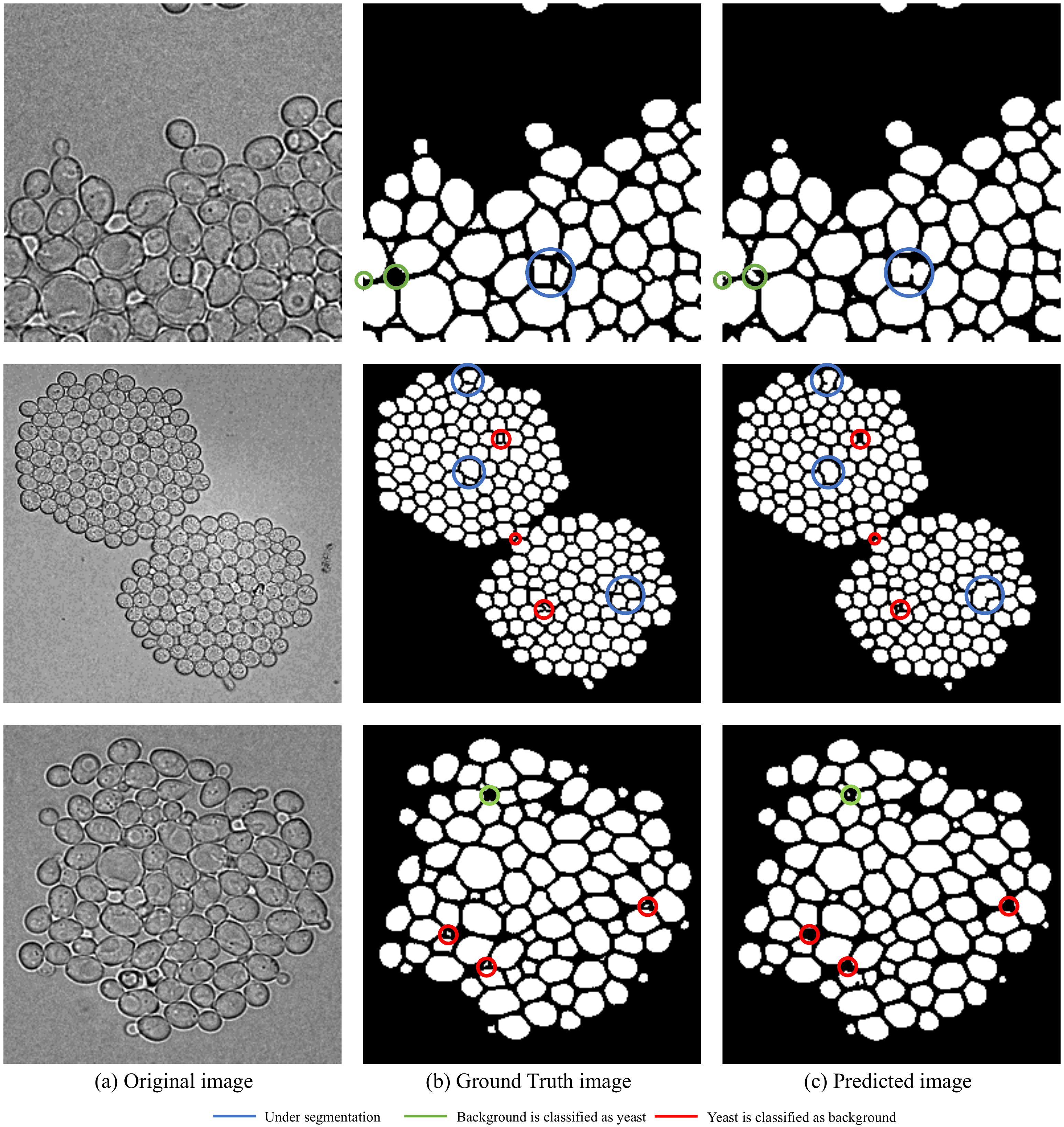}
\caption{An example of incorrect segmentation results using the proposed~PID-Net.}
\label{Fig:wrong}
\end{figure*}

Deep learning is essentially to build a probability distribution model driven by data. 
Therefore, as~the deep-learning-network architecture becomes deeper, the~quantity and quality of training data will have  a greater impact on the performance of the model. 
However, in~the imaging process of microorganism images, the~amount of satisfactory data is relatively small
due to some objective reasons, such as the impurities in the acquisition environment, uneven natural light and~other adverse factors, which leads to insufficient training and poor performance in various tasks.
Though the proposed PID-Net has excellent segmentation and counting performance for images with dense tiny objects, there still exists some mis-segmentation, causing the decrease of counting accuracy.
Several incorrect segmentation results are shown in Figure~\ref{Fig:wrong}.

There are three main problems for segmentation and counting, which are illustrated in Figure~\ref{Fig:wrong}.
The blue circle refers to the situation of under segmentation---that is, the~neighbor yeast cells cannot be segmented, and the edges cannot be detected.
Due to the counting method is based on the eight neighborhood search algorithm, the~situation leads to under estimation of the real count of yeast cells.
The green circle refers to a part of background is classified as yeast cells.
As shown in Figure~\ref{Fig:wrong}, most of the images are full of dense tiny yeast cells with irregular shapes, and~the limitation of small dataset leads to inadequate training. 

Therefore, the~background between yeast cells with irregular shape is easily classified as a yeast cell, which results in over estimation of the real count of yeast cells.
On the contrary, the~red circle represents the part of yeast cell is classified as background. 
There are 1 to 256 yeast cells with different sizes in a single image in this dataset, and thus the shape and size of yeast cells have a great difference.
Therefore, the~tiny yeast cell between the larger cells has a great similarity with the background, which is difficult for models to discriminate especially in a small dataset.
The situation leads to under estimation of the real count of yeast~cells.

The situations of adherent yeast cells and mis-segmentation lead to counting error.
Moreover, the~training data is limited due to the small dataset; therefore, the models cannot be trained perfectly.
The small dataset is a limitation of the yeast counting task.
However, despite  some cases of mis-segmentation, most of the yeast cells in the test dataset could be detected and segmented with other cells. The~segmented region might be small but~has little effect on the counting results calculated  using the eight neighborhood search~algorithm.

\section{Conclusions and Future~Work}\label{Sec:conclusion}

In this paper, a~CNN-based PID-Net was proposed for dense tiny objects (yeast) counting task. 
The PID-Net is an end-to-end model based on an encoder--decoder structure, and~we proposed a new down-sampling model consisting of pixel interval down-sampling and max-pooling, which can serve to extract the dense and sparse features in the task of dense tiny object counting. 
By comparing with the proposed PID-Net and classical U-Net-based yeast counting results, the~evaluation indices of Accuracy, Dice, Jaccard, Precision, Counting Accuracy and Hausdorff Distance of PID-Net were 97.51\%, 95.86\%, 92.10\%, 96.02\%, 96.97\% and 4.6272, which are improved by 0.04\%, 0.15\%, 0.26\%, 0.4\% and 5.7\%, respectively, and~the Hausdorff Distance  decreased by 0.0394.

Although the small image dataset resulted in some cases of mis-segmentation,~the proposed PID-Net showed a more satisfactory segmentation performance than the other models in the task of dense tiny object counting on a small~dataset.

In the future, we plan to apply PID-Net for more dense tiny object counting tasks, such as the \emph{streptococcus} counting task and blood-cell-counting task. 
We will  further optimize the PID-Net for better counting performance. For~instance,~object separation is one of the most significant parts in object counting; therefore, the~Contour Loss~\cite{Chen-2019-CLBL} can be used by referring to our work to distinguish inner texture and contour boundaries for more accurate counting. 
We also consider using Knowledge Distillation~\cite{Gou-2021-KDAS} to reduce the memory cost of PID-Net, which can help to deploy the model on  portable~equipment.

\section*{Acknowledgements}
This work is supported by the ``National Natural Science Foundation of China'' (No.61806047) 
and the ``Fundamental Research Funds for the Central Universities'' (No. N2019003). 
We thank Miss Zixian Li and Mr. Guoxian Li for their important discussion.

\section*{Declaration of competing interest} 
The authors declare that they have no conflict of interest.

\bibliographystyle{elsarticle-num}
\bibliography{wenxian}

\end{document}